\pdfoutput=1

\documentclass[11pt]{article}

\usepackage{acl}
\usepackage{times}
\usepackage{latexsym}
\usepackage{color}
\usepackage{arydshln}
\usepackage[T1]{fontenc}

\usepackage[utf8]{inputenc}

\usepackage{microtype}
\usepackage{graphics}
\usepackage{graphicx} 
\title{Generative Zero-Shot Prompt Learning for Cross-Domain Slot Filling with Inverse Prompting}

\author{Xuefeng Li$^{1}$\thanks{\ \ The first three authors contribute equally. Weiran Xu is the corresponding author.} ,
Liwen Wang$^{1*}$,
Guanting Dong$^{1*}$,\\
\bf Keqing He$^{2}$,
Jinzheng Zhao$^{3}$,
Hao Lei$^{1}$,
Jiachi Liu$^{1}$,
Weiran Xu$^{1}$\\ 
$^1$Beijing University of Posts and Telecommunications, Beijing, China\\
$^{2}$Meituan Group, Beijing, China\\
$^{3}$School of Computer Science and Electronic Engineering, University of Surrey, UK\\
\texttt{\{lixuefeng,w\_liwen,dongguanting,leihao,ljc1997\}@bupt.edu.cn}\\
\texttt{kqin@bupt.cn},
\texttt{j.zhao@surrey.ac.uk},
\texttt{xuweiran@bupt.edu.cn}
}

\begin{document}
\maketitle
\begin{abstract}

Zero-shot cross-domain slot filling aims to transfer knowledge from the labeled source domain to the unlabeled target domain. Existing models either encode slot descriptions and examples or design handcrafted question templates using heuristic rules, suffering from poor generalization capability or robustness. In this paper, we propose a generative zero-shot prompt learning framework for cross-domain slot filling, both improving generalization and robustness than previous work. Besides, we introduce a novel inverse prompting strategy to distinguish different slot types to avoid the multiple prediction problem, and an efficient prompt tuning strategy to boost higher performance by only training fewer prompt parameters. Experiments and analysis demonstrate the effectiveness of our proposed framework, especially huge improvements (+13.44\% F1) on the unseen slots.\footnote{Our source code is available at: \url{https://github.com/LiXuefeng2020ai/GZPL}}
\end{abstract}
\section{Introduction}


Slot filling in a task-oriented dialogue system aims to extract task-related information like \emph{hotel\_name}, \emph{hotel\_address} from user queries, which is widely applied to existing intelligent conversation applications \cite{Tulshan2019SurveyOV,zhang2020recent}. Traditional supervised methods \cite{Zhang2016AJM,goo-etal-2018-slot,qin-etal-2019-stack,wu2020slotrefine,He2020SyntacticGC,he-etal-2020-learning-tag} have shown remarkable performance, but they still rely on large-scale labeled data. Lack of generalization to new domains hinder its further application to practical industrial scenarios.

In this work, we focus on zero-shot cross-domain slot filling which transfers knowledge from the source domain $D_{S}$ to the target domain $D_{T}$ without requiring any labeled training data of $D_{T}$. Conventional approaches \cite{bapna2017towards,shah2019robust,he-etal-2020-contrastive,wang2021bridge} formulate slot filling as a sequence labeling task and use meta-information such as slot descriptions and slot examples to capture the semantic relationship between slot types and input tokens. However, these models only learn a surface mapping of the slot types between $D_{S}$ and $D_{T}$ and get poor performance on unseen slots in the target domain \cite{wang2021bridge}. Further, \cite{Lee2019ZeroShotAT,mehri2021gensf,du-etal-2021-qa,yu2021cross} propose a machine reading comprehension (MRC) framework for slot filling to enhance the semantic interaction between slot types and slot values. They firstly construct many well-designed question templates based on slot schema or slot examples, then train an MRC model \cite{rajpurkar-etal-2018-know} to predict corresponding slot values for a given slot type question. But they rely on handcrafted question templates using heuristic rules and pre-defined ontologies, which suffers from poor model robustness. Besides, employing additional pre-training on large-scale external MRC datasets is also time-consuming and prohibitively expensive.
\begin{figure}
    \centering
    \resizebox{.48\textwidth}{!}{
    \includegraphics{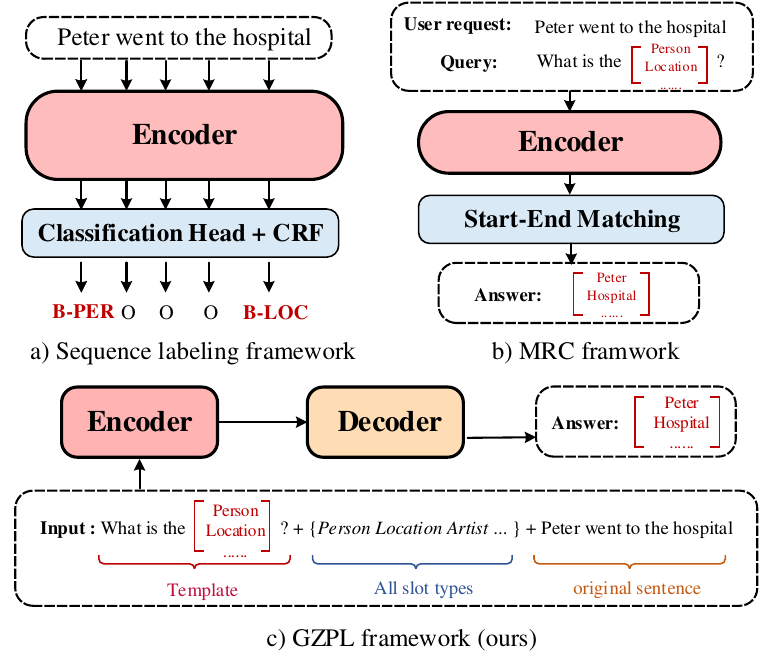}}
    \vspace{-0.75cm}
    \caption{Illustration of different frameworks for zero-shot slot filling.}
    \label{fig:intro}
    \vspace{-0.75cm}
\end{figure}

\begin{figure*}
    \centering
    \resizebox{.92\textwidth}{!}{
    \includegraphics{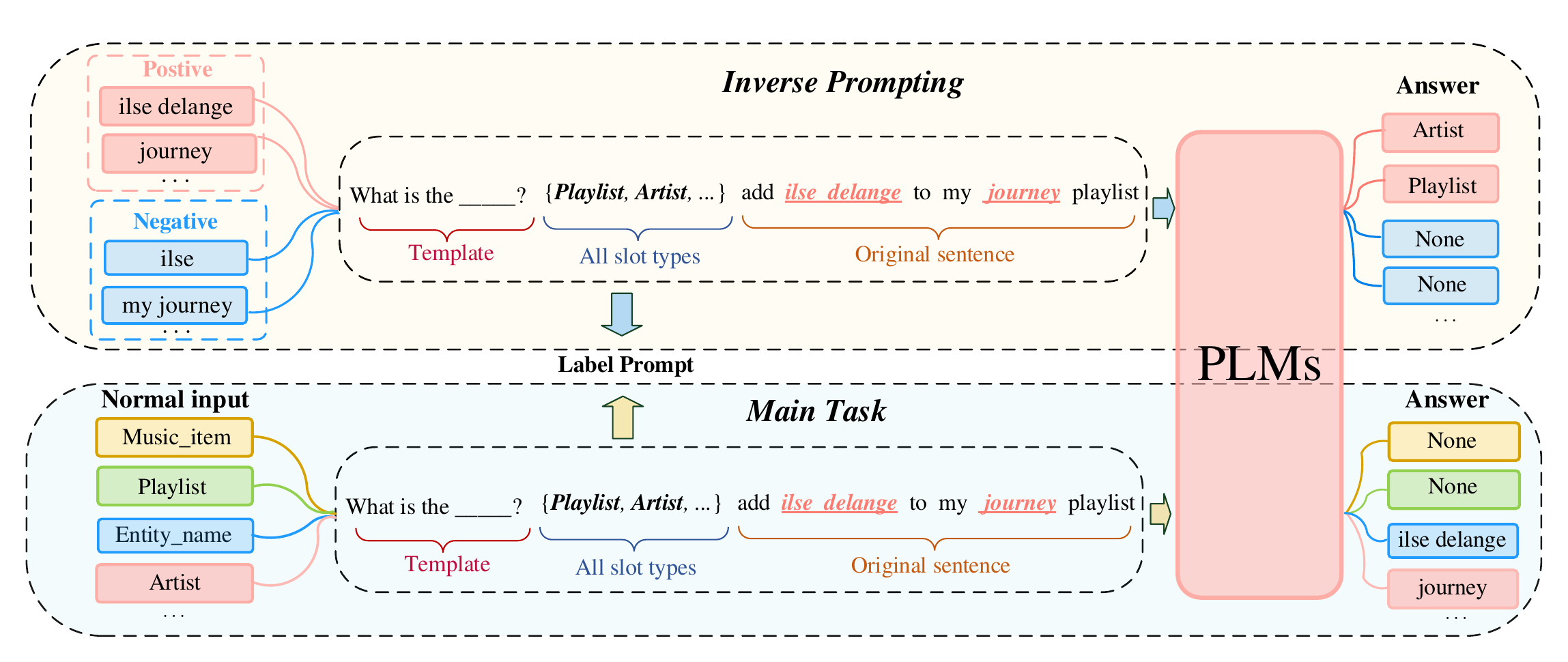}}
    \vspace{-0.5cm}
    \caption{The overall architecture of our proposed GZPL framework with inverse prompting.}
    \label{fig:model}
    \vspace{-0.6cm}
\end{figure*}

To solve the above issues, in this paper, we propose a \textbf{G}enerative \textbf{Z}ero-shot \textbf{P}rompt \textbf{L}earning (\textbf{GZPL}) framework for cross-domain slot filling. Instead of transforming the slot filling task into sequence labeling or MRC, we formulate it as a language generation task (see Fig \ref{fig:intro}). Specifically, we concat the question of each slot type, names of all slot types, the input query together to construct the input sequence and take the related slot values as output sequence. The converted text-to-text format has two benefits for zero-shot slot filling: (1) Compared to sequence labeling, our formulation enriches deep semantic interaction between slot types and slot values via pre-trained language models \cite{raffel2020exploring}, which helps recognize unseen slots only existing in the target domain. We find it significantly improves unseen slot F1 by 13.44\% compared to the previous state-of-the-art (SOTA) model (see Section \ref{unseen}). The result proves the strong generalization capability to new domains of our proposed framework. (2) Compared to MRC, our framework reduces the complexity of creating well-designed question templates and is more robust to different templates (see Section \ref{robustness}). Besides, we concat the names of all slot types into the input sequence to construct direct connections between different slot types, while MRC makes independent predictions for each slot type. Along with our proposed framework, we present an inverse prompting strategy to distinguish different slot types for a given entity to avoid the multiple prediction problem \cite{He2020ContrastiveZL} where the model possibly predicts multiple slot types for one entity span. Different from the above formulation, we take each slot value as input and corresponding slot type as output to build a mapping from entity tokens to entity types. In this way, we force the model to learn explicit distinctions of different types. Inspired by recent parameter-efficient tuning work \cite{li-liang-2021-prefix, lester-etal-2021-power}, we also introduce an efficient prompt tuning strategy to boost higher performance by training fewer prompt parameters instead of the whole PLM.


Our contributions are three-fold: (1) We propose a simple but strong generative zero-shot prompt learning framework for cross-domain slot filling, which has better generalization capability and robustness than previous work. (2) We present a novel inverse prompting strategy to distinguish different slot types to avoid the multiple prediction problem. Besides, we introduce an efficient prompt tuning strategy to boost higher performance only training fewer prompt parameters. (3) Experiments and analysis demonstrate the effectiveness of our proposed framework, especially for good generalization to unseen slots (F1 $+13.44\%\uparrow$), strong robustness to different templates ($\Delta$ F1 $+10.23\% \uparrow$), parameter efficiency (10x fewer parameters).

\section{Methodology}


Our model is shown in Fig \ref{fig:model}. In our framework, we first construct several simple template sentences for the model input, where each sentence includes a slot type question, all slot types and the original query. Then we use a PLM to generate the corresponding slot values. Along with the main task formulation, we perform an inverse-prompting task to warm up the parameters to strengthen the relationship between entities and slot types.



\subsection{Problem Definition}
Given a user input sentence containing $n$ words $\textbf{X}_{input} = \{x_1,x_2,...,x_n\}$  and slot type sets $\textbf{S} = \{s_1,s_2,...,s_m\}$,  the slot filling task aims to find all the entities in $\textbf{X}_{input}$. For zero-shot setting in our paper, we train models using labeled data from the source domain and make predictions in the target domain.


\subsection{Generative Zero-shot Prompt Learning Framework}
We customize the entire task using a generative zero-shot prompt learning framework. Specifically, we concat the question of each slot type, names of all slot types, the input query together to construct the input sequence and take the related slot values as output sequence. We formulate it as follows:
$$
\setlength{\abovedisplayskip}{3pt}
\setlength{\belowdisplayskip}{3pt}
   what\;is\;the \; \textbf{slot\_type}\;?\;\textbf{\{all\;slot\;types\}}\;x_1\;x_2\;...\;x_n 
$$

 where $\textbf{{slot\_type}}$  represents the queried slot type, $\textbf{\{all\;slot\;types\}}$ represents all slot types across all domains. For slot types that do not exist in the input, we set the answer to special token "none". For each original input query, we construct QA pairs as the same number of slot types\footnote{Appendix \ref{formats details} shows more details about input and output formats. Appendix \ref{inverse-prompt task} gives the analysis of the inverse-prompting task.}. 

\textbf{Label Prompt Construction} We do not focus on the question template construction as the previous works \citet{du-etal-2021-qa,yu2021cross}. Instead, we simply set up the simplest question form of ``$what \; is \; the \; ?\ $" to highlight the simplicity and effectiveness of our proposed framework. It is worth noting that we also include slot names from all domains in the prompt. The main purpose of this setting is to enhance the interaction between different slot types, so that the model can find the best answer from the original text.
\textbf{Inverse Prompting}
Previous MRC works suffer from the multiple prediction problem
\cite{He2020ContrastiveZL} where the model possibly predicts multiple slot types for one entity span. To solve such conflict, we design an invert prompting task to warm up the model parameters first. We inverse the original QA pair, that is, set the question to the entities and the answer to the corresponding slot types. This task enables the model to distinguish different slot types for slot entities. In this way, deep semantic relationships between slot types are learned, and the model will learn stronger entity-slot relations. We both train the main task and the inverse task in the same auto-regressive way. Experiments show that first using the inverse task for pre-training then the main task gets the best performance.


In addition, since the result of the main task could be "none", we additionally use a negative sampling strategy here to ensure the consistency of the two tasks. We just randomly sample different spans in sentences, and set the corresponding answers to "none". This strategy can also improve the anti-noise ability of the model and improve the robustness of the framework. In our experiments, we set the ratio of positive and negative samples to 1:1.

\textbf{Training and Inference}
During training, we try two different training strategies: fine-tuning and prefix-tuning \cite{li2021prefix}. In the  fine-tuning mode, we first use the inverse task to warm up the model parameters, and then perform the main task. All the PLM parameters are finetuned. For prefix-tuning, the parameters of the pre-trained model are fixed during training, and only the parameters of the new added prefix embeddings are trained. Specifically, we add a trainable prefix embedding matrix in each attention layer of the PLM \footnote{Please see more details in the original prefix-tuning work \cite{li2021prefix}.}. This method requires 10x fewer trainable parameters and is more parameter-efficient. 

During the inference, we only perform the main task. We query for all slot types, and the model directly generates the corresponding slot entities. Compared with the previous method \cite{yu2021cross}, our model will not need additional span matching mechanism, so it will be more concise and intuitive. To ensure task consistency with MRC-based models, we add a post-processing step: if multiple slot types predict the same entity span, we choose the answer with the highest generation probability of the first word.
\begin{table*}[t]
\centering
\resizebox{.9\textwidth}{!}{
\begin{tabular}{c|ccccc|cc|cc:cc}  
\hline
Training Setting & \multicolumn{5}{c|}{Sequence tagging-based models} & \multicolumn{2}{c|}{MRC-based models}& \multicolumn{4}{c}{Our models}\\
\hline
Domain $\downarrow$ $\sim$ Model $\rightarrow$ & CT & RZT & Coach & CZSL & PCLC & QASF & RCSF* & GZPL(ft) & GZPL(pt)& GZPL*(ft) & GZPL*(pt)  \\
\hline
AddToPlaylist & 38.82& 42.77& 50.90 & 53.89 & 59.24 & 59.29 & \textbf{68.70} &57.52 & \textbf{59.34} & 59.83& 61.64\\
BookRestaurant & 27.54& 30.68& 34.01 & 34.06& 41.36& 43.13& \textbf{63.49}& 57.50& \textbf{63.77} & 61.23& 62.93 \\
GetWeather & 46.45 & 50.28 & 50.47 & 52.04 & 54.21& 59.02& \textbf{65.36} & \textbf{64.90}& 64.20 & 62.58& 64.97\\
PlayMusic & 32.86 & 33.12 & 32.01 &  34.59& 34.95& 33.62& 53.51& 54.35 & \textbf{56.78}  & 62.73& \textbf{66.42} \\
RateBook & 14.54 & 16.43 & 22.06 &  31.53& 29.31& 33.34& 36.51& 31.86& \textbf{38.88}  & 45.88& \textbf{47.53} \\
SearchCreativeWork & 39.79 & 44.45 & 46.65 & 50.61& 53.51& 59.90& 69.22& 66.97& \textbf{71.96}  & 71.30& \textbf{72.88}\\
SearchScreeningEvent & 13.83 & 12.25 & 25.63  & 30.05 & 27.17& 22.83& 33.54& 44.80& \textbf{49.83} & 48.26& \textbf{51.42} \\
\hline
Average F1 & 30.55 & 32.85 & 37.39 & 40.99 & 42.82& 44.45 & 55.76& 53.99& \textbf{57.82} &58.82& \textbf{61.07} \\
\hline
\end{tabular}
}
\vspace{-0.3cm}
\caption{Slot F1-scores (\%) on SNIPS for different target domains under zero-shot settings. ft and pt stands for fine-tuning and prefix-tuning respectively. * indicates the backbone model is a large version of pre-trained model.}
\label{tbl:control-results1}
\vspace{-0.5cm} 
\end{table*}
\section{Settings}

\subsection{Datasets}


SNIPS \cite{coucke2018snips} is a public spoken language understanding dataset consisting of crowdsourced user utterances with 39 slots across 7 domains. It has around 2000 training instances per domain. To simulate the cross-domain scenarios, we follow \citet{liu2020coach} to split the dataset, which selects one domain as the target domain and the other six domains as the source domains each time. 

\subsection{Baselines}
\label{baselines}
Sequence Tagging Models: \textbf{Concept Tagger (CT)} proposed by \cite{bapna2017towards}, which utilizes slot descriptions to boost the performance on detecting unseen slots. \textbf{Robust Zero-shot Tagger (RZT)} proposed by \cite{shah2019robust}, which is based on CT and leverages both slot descriptions and examples to improve the robustness of zero-shot slot filling. \textbf{Coarse-to-fine Approach (Coach)} proposed by \cite{liu2020coach}, which contains coarse-grained BIO 3-way classification and a fine-grained slot type prediction. In this model, slot descriptions are used in the second stage to help recognize unseen slots, and template regularization is applied to further improve the slot filling performance of similar or the same slot types. \textbf{Contrastive Zero-Shot Learning with Adversarial Attack (CZSL-Adv)} proposed by \cite{he-etal-2020-contrastive}, which is based on Coach and utilizes contrastive learning and adversarial attacks to improve the performance and robustness of the framework. \textbf{Prototypical Contrastive Learning and Label Confusion (PCLC)} \cite{wang2021bridge}, which proposes a method to dynamically refine slot prototypes’ representations based on Coach framework and obtains an improved performance.

MRC-based Models: \textbf{QA-driven Slot Filling Framework (QASF)}. Contrary to previous methods, \citet{du-etal-2021-qa} introduced MRC-based framework and leveraged the PLMs to solve the problem. \textbf{Reading Comprehension for Slot Filling (RCSF)} \cite{yu2021cross}, which takes a new perspective on cross-domain slot filling by formulating it as a machine reading comprehension (MRC) problem, which transforms slot names into well-designed queries to improve the detection performance of domain-specific slots.

\subsection{Implementation Details}
\label{details}
We use T5-base\footnote{T5 is a transformer-based pre-training language model, whose pre-training tasks include text-to-text formulation. We select it as our pre-training model for the consistency between the pre-training tasks and the downstream slot-QA tasks.} as the backbone in our experiments. Model parameters are optimized using the AdamW optimizer \cite{kingma2014adam} with a learning rate 5e-05. We set the batch size to 8 and use early stop with a patience 10 to ensure the stability of the model. The prefix length is set to 5 and the dropout rate is set to 0.1. Since RCSF uses the BERT-Large\footnote{https://huggingface.co/deepset/bert-large-uncased-whole-word-masking-squad2
} model, we use T5-large\footnote{https://huggingface.co/t5-large} model to match the number of parameters of the model used in RCSF. The number of parameters of T5-base\footnote{https://huggingface.co/t5-base}, T5-large and prefix parameters are 2.2 billion, 7.7 billion, and 20 million, respectively. For all experiments, we train and test our model on 3090 GPU and use f1-score as the evaluation metric. During the training process, we only do prefix-tuning on T5-base, we fix the parameters of T5-base and only fine-tune the parameters of prefix embeddings. We take the average F1 scores of three experiments as our final result.




\section{Experiments}
\subsection{Main Results}
 Results show that our proposed framework GZPL significantly outperforms SOTAs. Our base model GZPL(pt) outperforms PCLC by 15.00\% and QASF by 13.37\% respectively. We don't directly compare our model with RCSF because it uses two unfair settings: using BERT-large as backbone and pre-training it on the QA dataset SQuAD2.0 \cite{rajpurkar2018know}. Nevertheless, our base model still outperforms RCSF by 2.06\%. We adopt another setting to compare with RCSF, that is, change the backbone model to T5-large to ensure that the model size is consistent. We can see GZPL*(pt) with T5-large outperforms RCSF by 6.31\%. Besides, we also find using prefix-tuning is better than traditional fine-tuning, which proves prefix-tuning has better knowledge transferability.\footnote{GZPL without special annotations represent using prefix-tuning unless otherwise noted in the following section.}

\begin{table}[t]\tiny
\centering
\resizebox{76mm}{6mm}{
\begin{tabular}{c|ccccc|c}  
\hline
& CT & RZT& Coach & PCLC& RCSF & GZPL  \\
\hline
seen & 37.23 & 40.99 & 46.22 & 51.69 &\textbf{75.96}& 66.49 \\

unseen & 3.38 & 2.19 & 9.31 & 17.38 & 26.21 & \textbf{39.65} \\
\hline
\end{tabular}}
\vspace{-0.3cm} 
\caption{Average F1 scores on seen and unseen slots across all target domains.}
\label{tbl:control-results2}
\vspace{-0.25cm} 
\end{table}

\begin{table}[t]\tiny
\centering
\resizebox{76mm}{7mm}{
\begin{tabular}{c|cccc}  
\hline
$\Delta$ F1 & del "what" & del "what is"& del "what is the"\\
\hline
GZPL & 2.4↓ & 3.0↓ & 7.2↓\\

RCSF & 12.8↓ & 14.1↓ & 19.8↓\\
\hline
\end{tabular}}
\vspace{-0.3cm} 
\caption{Average F1 score drop across all domains after the template changes. The smaller number indicates the better effect.}
\label{tbl:control-results4}
\end{table}

\begin{table}[t]
\centering
\resizebox{76mm}{5mm}{
\begin{tabular}{c|c c c c c c}
\hline
& 20 (1$\%$) & 50 (2.5$\%$) & 100 (5$\%$) & 200 (10$\%$) & 500 (25$\%$) & 2000 (100$\%$)\\
 \hline
 RCSF & 0.4 & 0.9 & 2.8 & 9.8 & 17.2 & 55.8\\
 GZPL & 6.2 & 23.7 & 36.5 & 41.5 & 48.2 & 57.8\\
\hline
\end{tabular}}
\caption{Averaged F1-scores (\%) over all target domains on SNIPS under the few-shot settings on source domains.}
\label{Table4}

\end{table}

\begin{table}[t]\tiny

\resizebox{78mm}{5mm}{
\begin{tabular}{c|c|c|c|c}  
\hline
 & GZPL& w/o LP &  w/o RP &  w/o (LP \& RP )   \\
\hline
Average F1 & 57.82 & 55.47 & 54.72 & 53.13 \\
\hline
\end{tabular}}
\vspace{-0.3cm}
\caption{Ablation studies. LP and RP stands for label prompt and inverse prompt, respectively. }
\label{tbl:control-results3}
\end{table}

\subsection{Analysis}
\label{unseen}
\textbf{Generalization Analysis}
Following \citet{wang2021bridge}, if a slot does not exist in the remaining six source domains, it will be categorized into the ``unseen slot" part, otherwise ``seen slot". The results are shown in Table \ref{tbl:control-results2}. We can see that 
our method outperforms previous methods by a large margin on unseen slots, while performs slightly worse than RCSF on seen slots. Our model focuses more on the generalizable knowledge transfer rather than overfitting on the seen slots in source domains, so it has stronger generalization ability than the previous methods.

\textbf{Robustness Analysis}
\label{robustness}
To verify the robustness of our framework, we change the original template "what is the ?" as RCSF. We still use the complete template during training, but delete some tokens of the template during testing, and the results are shown in Table \ref{tbl:control-results4}. Our model drops slightly by average 4.2\% when the template changes, while RCSF drops significantly by 15.6\%. This demonstrates that our model is more robust to different input templates.

\textbf{Effectiveness Analysis}
To further explore the effectiveness of the GZPL
under low resource scenarios, we conduct several
low-resource settings on source domains, which
means only 20, 50, 100, 200 and 500 samples
in source domain are used during training stage.
As SOTA model (RCSF) does not show results of
few-shot experiments, we evaluate RCSF using its
open source code. As shown in Table \ref{Table4}, the per
formance of our model is much better than that
of RCSF under low resource conditions. Besides,
with only 100 samples (5\%), our model maintains
63.13\% performance compared to the results using
complete source domain data. While using 500
samples (25\%), 82.08\% performance can be maintained. This demonstrates our approach is more
data-efficient than other slot filling models.

\textbf{Ablation Studies}
To better prove the effectiveness of the label prompt strategy and the inverse-prompt task, we conduct ablation experiments on these two components. Table \ref{tbl:control-results3} illustrates the results of ablation, where ``w/o" denotes the model performance without specific module. As we can see, the model will have a slight performance drop (-2.35$\%$) if the slot types in template are removed and the performance of the model will degrade significantly (-3.5$\%$) without the inverse-prompt task. Besides, it is observed that when removing both the label-prompt and inverse-prompt jointly, the performance of the model will drop drastically (-4.69$\%$). This suggests that both of them play an important role in improving the performance.


\section{Conclusion}
In this paper, we introduce a generative prompt learning framework for zero-shot cross-domain slot filling. Based on this, we introduce the label prompt strategy and the inverse prompting to improve the generalization capability and robustness of the framework. Another prefix-tuning mechanism is performed to boost model training efficiency. The exhaustive experimental results show the effectiveness of our methods, and the qualitative analysis inspire new insight into related area. Generally, our framework can be applied to more complex situations, such as nested NER, discontinuous/multiple slots, which we leave to future work. Another interesting direction is to improve the inference efficiency, like concat all the slot questions together and get final results.

\section{Acknowledgements}
This work was partially supported by National Key R\&D Program of China No. 2019YFF0303300 and Subject II  No. 2019YFF0303302, DOCOMO Beijing Communications Laboratories Co., Ltd, MoE-CMCC "Artifical Intelligence" Project No. MCM20190701.
\bibliography{anthology,custom}

\appendix




\section{Details about the input and output formats}
\label{formats details}
Table \ref{tab: example details} shows an example of how to perform slot filling tasks for a user query under our settings. As shown in the table, since we already know the slot type information for the domain the data belongs to, we will customize the unique questions for each slot type according to our template and the model then generate the answers for each question. The answer can be one or more spans in the original sentence, or be the special token "none". It is worth noting that when a slot type corresponds to multiple slot entities, the answer will be separated by commas. However, this situation hardly exists in the Snips dataset, so it is rare to have multiple spans as answers when testing.


  


\section{Analysis of the Inverse-prompting Task}
\label{inverse-prompt task}
\begin{figure}[t]
    \centering
    \centerline{
    \includegraphics[width=9cm]{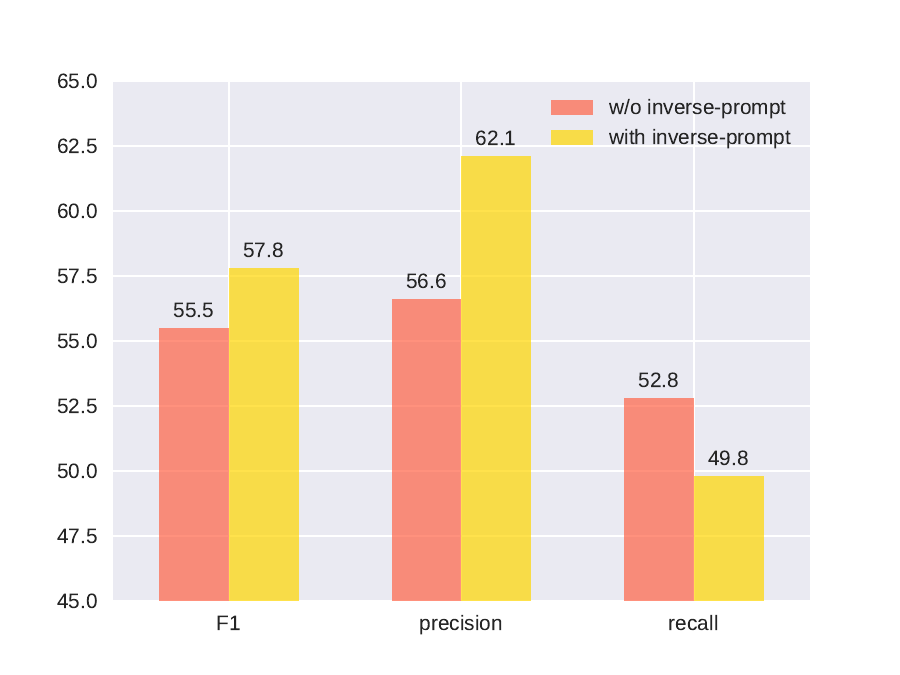}}
    \vspace{-0.2cm}
    \caption{Impact of the proposed inverse-prompt task on F1, precision and recall scores.}
    \label{fig:5}
\vspace{-0.5cm}
\end{figure}

To further explore whether our auxiliary task alleviates the problem of repeated generation, we verify its effect through the following two metrics: precision and recall score. We use these metrics based on our recognition that repeated generation will result in more entities being predicted. On the one hand, this will improve the recall score, and on the other hand, it will hurt the accuracy of the model prediction. The experimental results are shown in Figure \ref{fig:5}. As can be seen from the figure, after adding this inverse-prompt task, the recall-score of the model decreased by 3\%, while the precision-score increased by 5.5\%, which also increased the overall f1-score by 2.4\%. We also conducted a case study on the output of the model, and the results are shown in Table \ref{tab:data example}. After the tasks are added, the repeated generation of the model is significantly reduced. These results above illustrate that the proposed task enables the model to learn deep relationships between slot types, thereby reducing the problem of repeated generation.

\section{Limitations and Future Work}
The current work does achieve better performance than previous methods, but processing only one slot type at a time also reduces the efficiency of the model. In the future, we will explore how to maximize model efficiency. It would be an interesting challenge to generate answers for all the slots at once without degrading the effect of the model. Also, we will also try to apply our framework to more scenarios, such as NER and other tasks to explore the adaptability of the proposed method.

\begin{table*}[htbp]
\centering
\resizebox{.95\textwidth}{!}{
\begin{tabular}{c|c}
\hline
\textbf{Domain} & \textbf{SearchCreativeWork} \\
 \hline
slot types in this domain & object type, object name \\
\hline
all\_alot\_types & artist, playlist....object type, object name....\\
\hline
query & play the \textcolor{red}{game} \textcolor{green}{sugarfoot} \\
\hline
input1 &what is the object type ?  artist, playlist....object type, object name.... play the \textcolor{red}{game} sugarfoot\\
\hline
output1 & \textcolor{red}{game} \\
\hline
input2 & what is the object name ?  artist, playlist....object type, object name.... play the game \textcolor{green}{sugarfoot} \\
\hline
output2 & \textcolor{green}{sugarfoot} \\
\hline
\end{tabular}}

\caption{An example showing the details of the input and output formats under our settings.}
\label{tab: example details}

\end{table*}

\begin{table*}[htbp]
\centering
\resizebox{.95\textwidth}{!}{
\begin{tabular}{c|c}
\hline
\textbf{Case Study} & \textbf{Data} \\
 \hline
Query & add ilse delange to my journey playlist \\
\hline
Answer & music\_item→none; playlist\_owner→none; entity\_name→none;  playlist→journey; artist→ilse delange\\
\hline
w/o Inverse Prompting & \textcolor{red}{\textbf{music\_item→ilse delange;}} playlist\_owner→none; entity\_name→none;  playlist→journey; artist→ilse delange\\
\hline
w Inverse Prompting &\textbf{ music\_item→none;} playlist\_owner→none; entity\_name→none;  playlist→journey; artist→ilse delange\\
\hline
\end{tabular}}

\caption{The case study of GZPL w/o Inverse Prompting}
\label{tab:data example}

\end{table*}

\label{sec:appendix}



\end{document}